\documentclass{article}

\usepackage[preprint]{neurips_2022}

\usepackage{amsmath,amsfonts,bm}

\def\eqref#1{equation~\ref{#1}}

\def\1{\bm{1}}

\def\vtheta{{\bm{\theta}}}

\DeclareMathAlphabet{\mathsfit}{\encodingdefault}{\sfdefault}{m}{sl}
\SetMathAlphabet{\mathsfit}{bold}{\encodingdefault}{\sfdefault}{bx}{n}

\newcommand{\R}{\mathbb{R}}

\DeclareMathOperator*{\argmax}{arg\,max}

\usepackage[utf8]{inputenc} 
\usepackage[T1]{fontenc}    
\usepackage{hyperref}       
\usepackage{url}            
\usepackage{booktabs}       
\usepackage{amsfonts}       
\usepackage{nicefrac}       
\usepackage{microtype}      
\usepackage{xcolor}         
\usepackage{algorithm}
\usepackage{algpseudocode}
\usepackage{rotating}  
\usepackage{lscape}
\usepackage{hyperref}
\usepackage{url}
\usepackage{wrapfig}
\usepackage{caption}
\usepackage{subcaption}
\usepackage{tabularx}
\usepackage{booktabs}

\usepackage{xfrac}

\DeclareMathOperator*{\ev}{\mathbb{E}}

\newcommand{\de}{\,\mathrm{d}}
\newcommand{\vrho}{\mathbr{\rho}}

\newcommand{\hyscoreprime}[1][\vtheta]{\nabla_{\vrho'}\log\nu_{\vrho'}}

\definecolor{citrine}{rgb}{0.89, 0.82, 0.04}
\definecolor{blued}{RGB}{70,197,221}

\usepackage{amsthm}  
\usepackage{thmtools}
\usepackage{thm-restate}

\definecolor{blue}{rgb}{0.122,0.467,0.706}
\definecolor{green}{rgb}{0.173,0.627,0.173}
\definecolor{orange}{rgb}{1.0,0.498,0.055}
\definecolor{red}{rgb}{0.893,0.153,0.157}

\usepackage[utf8]{inputenc} 
\usepackage[T1]{fontenc}    
\usepackage{hyperref}       
\usepackage{url}            
\usepackage{booktabs}       
\usepackage{amsfonts}       
\usepackage{nicefrac}       
\usepackage{microtype}      
\usepackage{xcolor}         

\newcommand{\approach}{GoGePo}

\usepackage{multibib}
\newcites{appendix}{Additional References}

\title{Goal-Conditioned Generators of Deep Policies}
\author{%
  Francesco Faccio\textsuperscript{\rm 1}\thanks{Equal contribution. Correspondence to francesco@idsia.ch}\\
  \And
  Vincent Herrmann\textsuperscript{\rm 1}\textsuperscript{$*$}\\
  \And
  Aditya Ramesh\textsuperscript{\rm 1}\\
  \And
  Louis Kirsch\textsuperscript{\rm 1}\\
  \And
  J{\"{u}}rgen Schmidhuber\textsuperscript{\rm 1,2,3}
  \AND
  {\normalfont \textsuperscript{\rm 1} The Swiss AI Lab IDSIA/USI/SUPSI} \\
    Lugano, Ticino, Switzerland\\
    \textsuperscript{\rm 2} AI Initiative, KAUST, Thuwal, Saudi Arabia\\
    \textsuperscript{\rm 3} NNAISENSE, Lugano, Switzerland\\
    }

\begin{document}

\maketitle

\begin{abstract}
Goal-conditioned Reinforcement Learning (RL) aims at learning optimal policies, given goals encoded in special command inputs. Here we study goal-conditioned neural nets (NNs) that learn to generate deep NN policies in form of context-specific weight matrices, similar to Fast Weight Programmers and other methods from the 1990s. Using context commands of the form ``generate a policy that achieves a desired expected return,'' our NN generators combine powerful exploration of parameter space with generalization across commands to iteratively find better and better policies. 
A form of weight-sharing HyperNetworks and policy embeddings scales our method to generate deep NNs. Experiments show how a single learned policy generator can produce policies that achieve any return seen during training. Finally, we evaluate our algorithm on a set of continuous control tasks where it exhibits competitive performance. Our code is public.\footnote{\url{https://github.com/IDSIA/GoGePo}}
\end{abstract}

\section{Introduction}
\label{sec:introduction}
General reinforcement learning (RL) is about training agents to execute action sequences that maximize cumulative rewards in possibly non-continuous, non-differentiable, partially observable environments~\citep{Kaelbling:96,hasselt2012,Schmidhuber:90sandiego}.
 Goal-conditioned RL agents can learn to solve many different tasks, where the present task is encoded by special command inputs~\citep{SchmidhuberHuber:91,Schaul:2015:UVF:3045118.3045258}. 

Many RL methods learn value functions~\citep{Sutton:2018:RLI:3312046} or estimate  stochastic policy gradients (with possibly high variance)~\citep{williams1992simple, Sutton1999}.
Upside-down RL (UDRL)~\citep{srivastava2019training, schmidhuber2019reinforcement} 
and related methods~\citep{ghosh2020learning}, however,
use supervised learning to train goal-conditioned RL agents. UDRL agents receive command inputs of the form "act in the environment and achieve a desired return within so much time"~\citep{schmidhuber2019reinforcement}.  Typically, hindsight learning~\citep{andrychowicz2017hindsight,rauber2018hindsight} is used to transform the RL problem into the problem of predicting actions, given reward commands.
This is quite powerful. Consider a command-based agent interacting with an environment, given a random command $c$, and achieving return $r$. Its behavior would have been optimal if the command had been $r$.
Hence the agent's parameters can be learned by maximizing the likelihood of the agent's behavior, given command $r$.
Unfortunately, in the episodic setting, many behaviors may satisfy the same command. Hence the function to be learned may be highly multimodal, and a simple maximum likelihood approach may fail to capture the variability in the data.\footnote{Note that in stochastic environments with episodic resets, certain UDRL variants will fail to maximize the probability of satisfying their commands~\citep{vstrupl2022upside}.}

To overcome this limitation, we introduce \emph{\approach}, a novel method for return-conditioned generation of policies evaluated in parameter space.
First, we use a Fast Weight Programmer (FWP)~\citep{schmidhuber1992learning,schmidhuber1993self,Ha2016} to generate the parameters of a desired policy, given a ``desired return'' command.
Then, we evaluate the policy using a parameter-based value function~\citep{harb2020policy,faccio2020parameter}.
This allows for end-to-end optimization of the return-conditioned generator producing deep NN policies by matching the commands (desired returns) to the evaluated returns.

The paper is structured as follows:
Section~\ref{sec:background} introduces the MDP frameworks for action-based and parameter-based methods;
Section~\ref{sec:fw} reviews the concept of Fast Weight Programmers;
Section~\ref{sec:cm} describes {\approach} including architectural choices;
Section~\ref{sec:exp} evaluates our method on continuous control tasks where it demonstrates competitive performance. Our analysis shows how a single learned policy generator can produce policies yielding any desired return seen during training.
Finally, we discuss related and future work in Sections~\ref{sec:related_work} and \ref{sec:conclusion}.
\section{Background}
\label{sec:background}
We consider a Markov Decision Process (MDP)~\citep{stratonovich1960,puterman2014markov} $\mathcal{M}=(\mathcal{S},\mathcal{A},P,R,\gamma,\mu_0)$. At each time step $t$, an artificial agent observes a state $s_t \in \mathcal{S}$, chooses an action $a_t \in \mathcal{A}$, obtains a reward $r_t = R(s_t,a_t)$ and transitions to a new state with probability $P(s_{t+1}| s_t, a_t)$. The initial state of the agent is chosen with probability $\mu_0$. The behavior of the agent is expressed through its stochastic policy $\pi_{\theta}: \mathcal{S} \rightarrow \Delta(\mathcal{A})$, where $\theta \in \Theta$ are the policy parameters. If for each state $s$ there is an action $a$ such that $\pi_{\theta}(a|s) = 1$, we will call the policy deterministic. The agent interacts with the environment through episodes, starting from the initial states, and ending either when the agent reaches a set of particular states---these can be failing states or goal states---or when it hits a time horizon $H$.
We define a trajectory $\tau \in \mathcal{T}$ as the sequence of state-action pairs that an agent encounters during an episode in the MDP $\tau = (s_{\tau,0}, a_{\tau,0}, s_{\tau,1}, a_{\tau,1}, \dots, s_{\tau,T}, a_{\tau,T})$, where $T$ denotes the time-step at the end of the episode ($T \leq H$). The return of a trajectory $R(\tau)$ is defined as the cumulative discounted sum of rewards over the trajectory $R(\tau) = \sum_{t=0}^{T}\gamma^t R(s_{\tau,t}, a_{\tau,t})$, where $\gamma \in (0,1]$ is the discount factor.

The RL problem consists in finding the policy $\pi_{\theta^*}$ that maximizes the expected return obtained from the environment, i.e. $\pi_{\theta^*} = \argmax_{\pi_{\theta}} J(\theta)$:
\begin{equation}
\label{eq:j_theta}
 J(\theta) = {\int}_{\mathcal{T}} p(\tau | \theta) R(\tau) \de \tau,    
\end{equation}
where $p(\tau | \theta) = \mu_0(s_0) \prod_{t=0}^T \pi_{\theta}(a_t|s_t) P(s_{t+1} | s_t, a_t)$ is the distribution over trajectories induced by $\pi_{\theta}$ in the MDP. When the policy is stochastic and differentiable, by taking the gradient of $J(\theta)$ with respect to the policy parameters we obtain an algorithm called REINFORCE~\citep{williams1992simple}: 
$ \nabla_{\theta} J(\theta) = {\int}_{\mathcal{T}} p(\tau | \theta) \nabla_{\theta} p(\tau | \theta) R(\tau) \de \tau$.

In parameter-based methods ~\citep{sehnke_parameterexploring_2010, sehnkepgpecontrol, salimans2017evolution, mania2018simple}, at the beginning of each episode, the weights of a policy are sampled from a distribution $\nu_{\rho}(\theta)$, called the hyperpolicy, which is parametrized by $\rho$. Typically, the stochasticity of the hyperpolicy is sufficient for exploration, and deterministic policies are used. The RL problem translates into finding the hyperpolicy parameters $\rho$ maximizing expected return, i.e. $\nu_{\rho^*} = \argmax_{\nu_{\rho}} J(\rho)$:
\begin{equation}
\label{eq:j_rho}
 J(\rho) = \int_{\Theta} \nu_{\rho}(\theta) {\int}_{\mathcal{T}} p(\tau | \theta) R(\tau) \de \tau \de \theta.    
\end{equation}
This objective is maximized by taking the gradient of $J(\rho)$ with respect to the hyperpolicy parameters: $ \nabla_{\rho} J(\rho) = \int_{\Theta} {\int}_{\mathcal{T}} \nu_{\rho}(\theta) \nabla_{\rho} \log \nu_{\rho}(\theta)  p(\tau | \theta) R(\tau) \de \tau \de \theta$. This gradient can be either approximated through samples~\citep{sehnke_parameterexploring_2010,sehnkepgpecontrol, salimans2017evolution} or estimated using finite difference methods~\citep{mania2018simple}. This only requires differentiability and stochasticity of the hyperpolicy.

For deterministic hyperpolicy and stochastic policy, the dependency on $\rho$ is lost and the policy parameters $\theta$ can be directly maximized using Equation~\ref{eq:j_theta}. Since the optimization problem is episodic, we can set the discount factor $\gamma$ to 1.

\section{Fast Weight Programmers}
\label{sec:fw}
Fast Weight Programmers (FWPs)~\citep{schmidhuber1992learning,schmidhuber1993self} are NNs that generate changes of weights of another NN conditioned on some contextual input.
In our UDRL-like case, the context is the desired return to be obtained by a generated policy.
The outputs of the FWP are the policy parameters $\theta \in \Theta$.
Formally, our FWP is a function $G_{\rho}: \mathbb{R}^{n_c} \longrightarrow \Theta$, where $c \in \mathbb{R}^{n_c}$ is the context-input and $\rho \in \mathrm{P}$ are the FWP parameters.
Here, we consider a probabilistic FWP of the form $g_{\rho}(\theta | c) = G_{\rho}(c) + \epsilon$, with $\epsilon \sim \mathcal{N}(0, \sigma^2 I)$ and $\sigma$ is fixed.
In this setting, the FWP conditioned on context $c$ induces a probability distribution over the parameter space, similar to the one induced by the hyperpolicy in Section~\ref{sec:background}.
Using the FWP to generate the weights of a policy, we can rewrite the RL objective, making it context-dependent:
\begin{equation}
\label{eq:j_rho_c}
     J(\rho, c) = \int_{\Theta} g_{\rho}(\theta | c) {\int}_{\mathcal{T}} p(\tau | \theta) R(\tau) \de \tau \de \theta.
\end{equation}
Compared to Eq.~\ref{eq:j_rho}, $J(\rho, c)$ induces a set of optimization problems that now are context-specific~\footnote{Note the generality of Eq.~\ref{eq:j_rho_c}.
In supervised learning, common FWP applications include the case where $g$ is deterministic, $\theta$ are the weights of an NN (possibly recurrent), $p(\tau|\theta)$ is the output of the NN given a batch of input data, $R(\tau)$ is the negative supervised loss.}.
Here, $J(\rho, c)$ is the expected return for generating a policy with a generator parametrized by $\rho$, when observing context $c$.
Instead of optimizing Eq.~\ref{eq:j_rho} using policy gradient methods, we are interested in learning a good policy through pure supervised learning by following a sequence of context-commands of the form ``generate a policy that achieves a desired expected return.''
Under such commands, for any $c$, the objective $J(\rho, c)$ can be optimized with respect to $\rho$ to equal $c$.
FWPs offer a suitable framework for this setting, since the generator network can learn to create weights of the policy network so that it achieves what the given context requires.

\section{Deep Policy Generators ({\approach})}
\label{sec:cm}
\begin{wrapfigure}{r}{0.5\textwidth}
\begin{center}
\centering
\includegraphics[width=\linewidth]{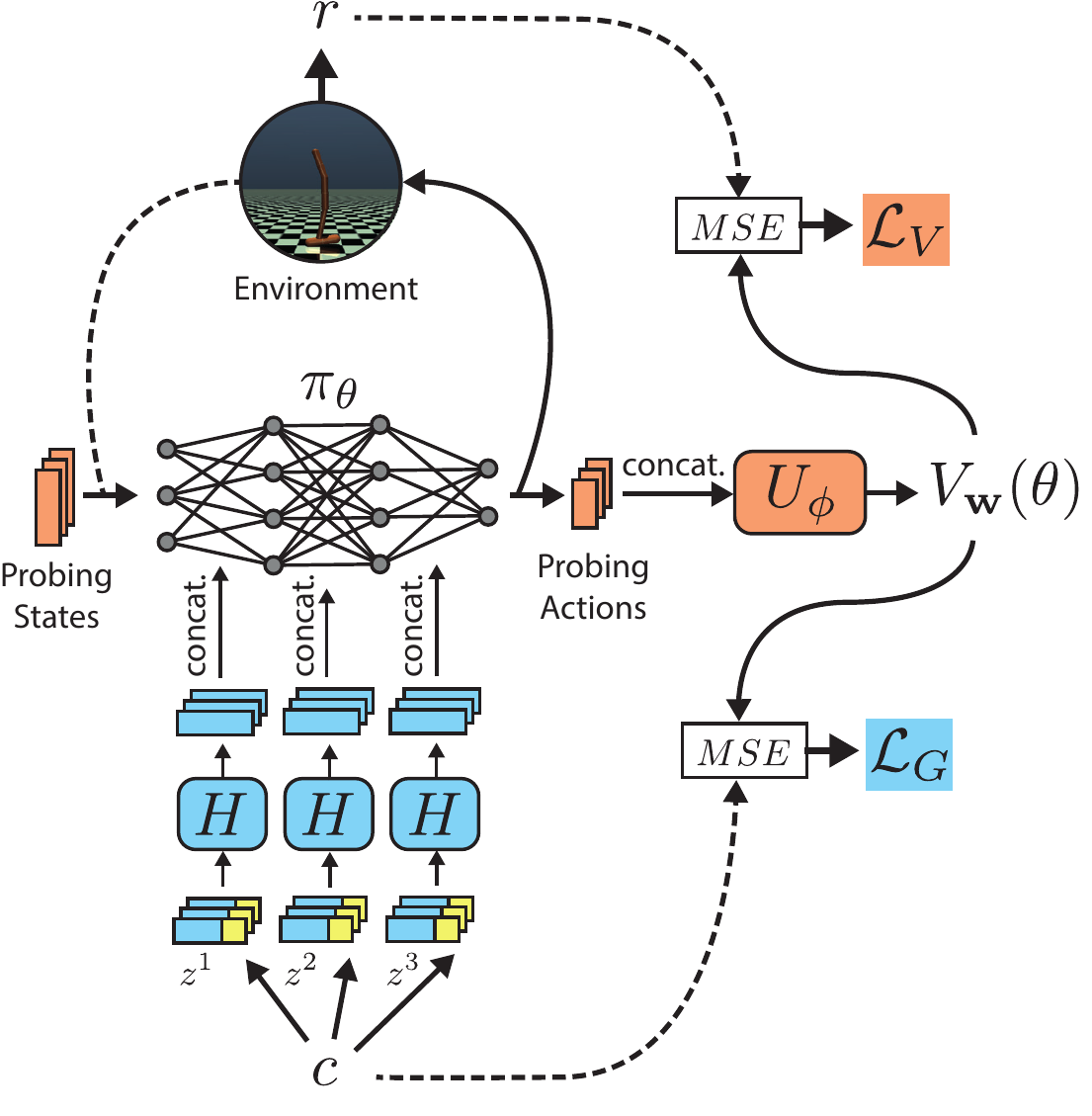}
\end{center}
\caption{
{\approach} generates policies using a Fast Weight Programmer (hypernetwork) conditioned on a desired return and evaluates the resulting policy using a parameter-based value function based on fingerprinting.
This enables training using supervised learning.
}
\label{fig:architecture}
\end{wrapfigure}

Here we develop \emph{\approach}, our algorithm to generate policies that achieve any desired return.
In the supervised learning scenario, it is straightforward to learn the parameters of the FWP that minimize the error $\mathcal{L}_G(\rho) =\ev_{c \in D, \theta \sim g_{\rho}(\cdot | c)} [(J(\theta) - c)^2]$, where the context $c$ comes from some set of possible commands $D$.
This is because in supervised learning $J(\theta)$, the expected return, is a differentiable function of the policy parameters, unlike in general RL.
Therefore, to make the objective differentiable, we learn an evaluator function $V_\textbf{w}: \Theta \longrightarrow \mathbb{R}$ parametrized by $\textbf{w}$ that estimates $J(\theta)$ using supervised learning~\citep{faccio2020parameter}.
This function is a map from the policy parameters to the expected return.
Once $V$ is learned, the objective $\mathcal{L}_G(\rho)$ can be optimized end-to-end, like in the supervised learning scenario, to directly learn the generator's parameters.
Concretely, we minimize $\mathcal{L}_G(\rho) =\ev_{c \in D} [(V_\textbf{w}(G_{\rho}(c)) - c)^2]$ to learn the parameters $\rho$.

Our method is described in Algorithm~\ref{alg:pvf} and consists of three steps.
\textbf{First}, in each iteration, a command $c$ is chosen following some  strategy.
Ideally, to ensure that the generated policies improve over time, the generator should be instructed to produce larger and larger returns.
We discuss command strategies in the next paragraph.
The generator observes $c$ and produces policy $\pi_{\theta}$ which is run in the environment.
The return and the policy $(r, \theta)$ are then stored in a replay buffer.
\textbf{Second}, the evaluator function is trained to predict the return of the policies observed during training.
This is achieved by minimizing MSE loss $\mathcal{L}_V(w) =\ev_{(r, \theta) \in B} [(r - V_{\textbf{w}}(\theta))^2]$. \textbf{Third}, we use the learned evaluator  to directly minimize $\mathcal{L}_G(\rho) = \ev_{r \in B} [(r - V_{\textbf{w}}(G_{\rho}(r)))^2]$.

\begin{algorithm}[H]
  \caption{{\approach} with return commands}
  \label{alg:pvf}
   \hspace*{\algorithmicindent} \textbf{Input}: Differentiable generator $G_{\rho}: \mathcal{R}  \rightarrow \Theta$ with parameters $\rho$;
   differentiable evaluator $V_{\textbf{w}}: \Theta \rightarrow \mathcal{R}$ with parameters $\textbf{w}$;
   empty replay buffer $D$\\
   \hspace*{\algorithmicindent} \textbf{Output} : Learned $V_{\textbf{w}} \approx V(\theta) \forall \theta$, learned $G_\rho \; s.t. \; V(G_\rho(r))\approx r \forall r$
    \begin{algorithmic}
    \State Initialize generator and critic weights $\rho, \textbf{w}$, set initial return command $c=0$
	\Repeat:
	    \State Sample policy parameters $\theta \sim g_\rho(\theta, c)$
		\State Generate an episode $s_{0}, a_{0}, r_{1}, s_{1}, a_{1}, r_{2}, \dots, s_{T-1}, a_{T-1}, r_{T}$ with policy $\pi_{\theta}$
		\State Compute return $r = \sum_{k=1}^T r_k$
		\State Store $(r, \theta)$ in the replay buffer $D$
		\For {many steps}:
		    \State Sample a batch $B = \{(r, \theta)\}$ from $D$
		    \State Update evaluator by stochastic gradient descent: $\nabla_{\textbf{w}} \ev_{(r, \theta) \in B} [(r - V_{\textbf{w}}(\theta))^2]$
	    \EndFor
		\For {many steps}:
		    \State Sample a batch $B = \{r\}$ from $D$
		    \State Update generator by stochastic gradient descent: $\nabla_{\rho} \ev_{r \in B} [(r - V_{\textbf{w}}(G_{\rho}(r)))^2]$
	    \EndFor
	    \State Set next return command $c$ using some strategy
    \Until{convergence}
    \end{algorithmic}
\end{algorithm}

\paragraph{Choosing the command}
\label{par:comm}
The strategy of choosing the command $c$ before interacting with the environment is important.
Intuitively, asking the generator to produce low return policies will not necessarily help finding better policies.
On the other hand, asking for too much will produce policies that are out of distribution, given the training data, and the generator cannot be trusted to produce such values.
Hence it is reasonable to ask the generator to produce a return close to the highest one observed so far.
More on command strategies can be found in Section~\ref{sec:exp_results}.

\paragraph{Scaling to deep policies}
Both generating and evaluating the weights of a deep feedforward MLP-based policy is difficult for large policies. 
The sheer number of policy weights, as well as their lack of easily recognizable structure, requires special solutions for generator and evaluator. To scale FWPs to deep policies, we rely on the relaxed weight-sharing of hypernetworks \citep{Ha2016} for the generator, and on parameter-based value functions~\citep{faccio2020parameter} using a fingerprinting mechanism \citep{harb2020policy} for the evaluator. We discuss these two approaches in the next section.

\subsection{HyperNetworks}
\label{sec:hn}

The idea behind certain feed-forward FWPs called hypernetworks~\citep{Ha2016} is to split the parameters of the generated network $\theta$ into smaller slices $s_l$. 
A shared NN $H$ with parameters $\xi$ receives as input a learned embedding $z_l$ and outputs the slice $s_l$, i.e. $s_l = H_{\xi}(z_l)$ for each $l$.
Following \cite{von2020continual}, further context information can be given to $H$ in form of an additional conditioning input $c$, which can be either either scalar or vector-valued: $s_l = H_{\xi}(z_l, c)$. Then the weights are combined by concatenating all generated slices:
\begin{equation}
\label{eq:hn_cat_general}
\theta = 
\begin{bmatrix}
s_{1} & s_{2} & s_{3} & \dots \\
\end{bmatrix}.
\end{equation}
The splitting of  $\theta$ into slices and the choice of $H$  depend on the specific architecture of the generated policy. Here we are interested in generating MLP policies whose parameters $\theta$ consist of weight matrices $K^j$ with $j \in \{1, 2, \dots, n_K\}$, where $n_K$ is the policy's number of layers. 
We use an MLP $H_{\xi}$ to generate each slice of each weight matrix: the hypernetwork generator $G_{\rho}$ splits each weight matrix into slices $s_{mn}^j \in \R ^ {f \times f}$, where $j$ is the policy layer, and $m,n$ are indexes of the slice in weight matrix of layer $l$. For each of these slices, a small embedding vector $z_{mn}^j \in \R ^d$  is learned. 
Our network $H_{\xi}$ is an MLP, followed by a reshaping operation that turns a vector of size $f^2$ into an $f \times f$ matrix:
\begin{equation}
\label{eq:hn_slice}
\begin{split}
s_{mn}^{j} &= H_\xi(z_{mn}^j, c).
\end{split}
\end{equation}
The slices are then concatenated over two dimensions to obtain the full weight matrices:
\begin{equation}
\label{eq:hn_cat}
K^j = 
\begin{bmatrix}
s_{11}^{j} & s_{12}^{j} & \dots \\[6pt]
s_{21}^{j} & s_{22}^{j} & \\
\vdots &  & \ddots \\
\end{bmatrix}.
\end{equation}

The full hypernetwork generator $G_\rho$ consists of the shared network $H_\xi$, as well as all embeddings $z_{mn}^j$.
Its learnable parameters are $\rho = \{\xi, z_{mn}^j \forall m, n, j\}$.

\begin{wrapfigure}{rh}{0.4\textwidth}
\centering
\includegraphics[width=0.63\linewidth]{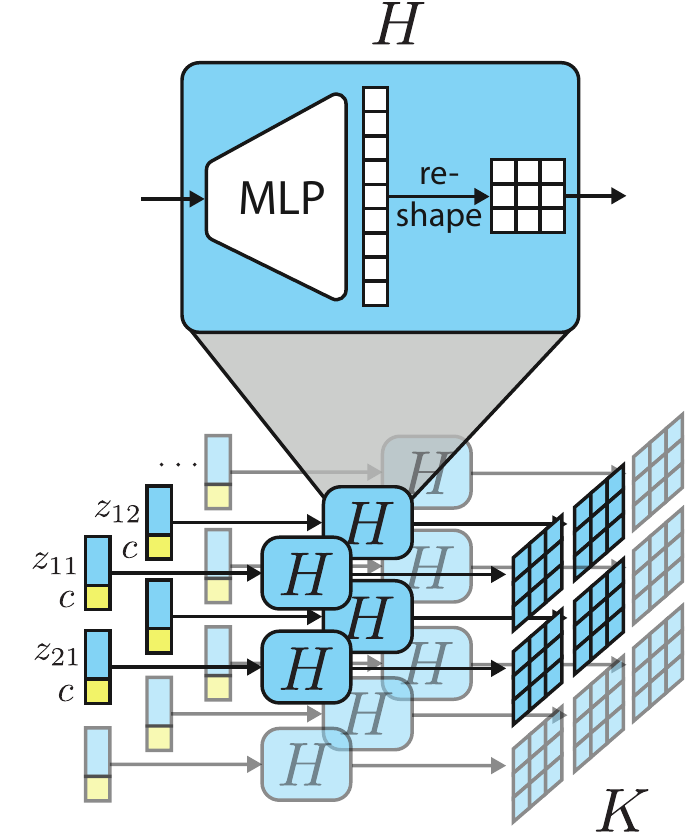}
\caption{Generating a weight matrix $K$ by concatenating slices that are generated from learned embeddings $z$ and return conditioning $r$ using a shared network $H$.}
\label{fig:hypernetwork}
\end{wrapfigure}

Generator $G_\rho$ is supposed to dynamically generate policy parameters, conditioned on the total return these policies should achieve. 
The conditioning input $c$ is simply this scalar return command.
It is appended to each learned slice embedding $z_{mn}^j$. 
The resulting vectors are the inputs to the network $H$.
Figure~\ref{fig:hypernetwork} shows a diagram of this process.

For the the slicing to work, the widths and heights of the weight matrices have to be multiples of $f$.
For the hidden layers of an MLP, this is easily achieved since we can freely choose the numbers of neurons.
For the input and output layers, however, we are constrained by the dimensions of environmental observations and actions.
To accommodate any number of input and output neurons, we use dedicated networks $H_i$ and $H_o$ for the input and output layers.
The generated slices have the shape $f \times n_i$ for the input layer ($n_i$ is the number of input neurons) and $n_o \times f$ for the output layer ($n_o$ is the number of output neurons).

\subsection{Policy Fingerprinting}
\label{sec:fing}
We use a policy fingerprinting mechanism~\citep{harb2020policy} as 
an effective method to evaluate the performance of multiple NNs through a single function.
Policy fingerprinting works by giving a set of learnable probing states as input to the policy $\pi_\theta$.
The resulting outputs of the policy---called probing actions---are concatenated and given as input to an MLP $U$ that computes the prediction $V_\textbf{w}(\theta)$. Here the set of parameters $\textbf{w}$ of this evaluator consists of the MLP parameters $\phi$ and all the parameters of the probing states. When training $V_\textbf{w}$, the probing states learn to query the policy in meaningful situations, so that the policy's success can be judged by its probing actions.
Fingerprinting is similar to a previous technique~\citep{learningtothink2015} where an NN learns to send queries (sequences of activation vectors) into another already trained NN, and learns to use the answers (sequences of activation vectors) to improve its own performance.
Figure~\ref{fig:architecture} shows a diagram of our method with a hypernetwork generator and a fingerprinting value function.

The benefits of policy fingerprinting over directly observing policy weights become apparent as soon as we have at least one hidden layer in an MLP policy: the weights then have a large number of symmetries, i.e., many different weight configurations that are entirely equivalent in terms of the input-output mapping of the network. 
The main symmetries reflect possible permutations of hidden neurons and scalings of the weight matrices \citep{kuurkova1994functionally}.

The probing actions of the fingerprinting mechanism are invariant with respect to such symmetries.
In fact, they are invariant even with respect to the general policy architecture.
This entails advantages not only for the value function $V_\textbf{w}$, but also for the generator:
the gradients w.r.t. the generator's weights $\rho$ are obtained by backpropagating through $V_\textbf{w}$.
If $V_\textbf{w}$ is fingerprinting-based, these gradients will point only in directions which, when followed, actually yield changes of the generated policy's probing actions.
Consequently, the generator will ignore potential policy weight changes that have no effect on the policy's probing actions (which are proxies for the policy's general behavior in the environment).

\section{Experiments}
\label{sec:exp}
We empirically evaluate {\approach} as follows:
First, we show competitive performance on common continuous control problems.
Then we use the the learned fingerprinting mechanism to visualize the policies created by the generator over the course of training, and investigate its learning behavior. The implementation of {\approach} will be made available with the publication.

\subsection{Results on continuous control RL environments}
\label{sec:exp_results}

\begin{figure}[t]
\begin{center}
\centering
\includegraphics[width=0.8\linewidth]{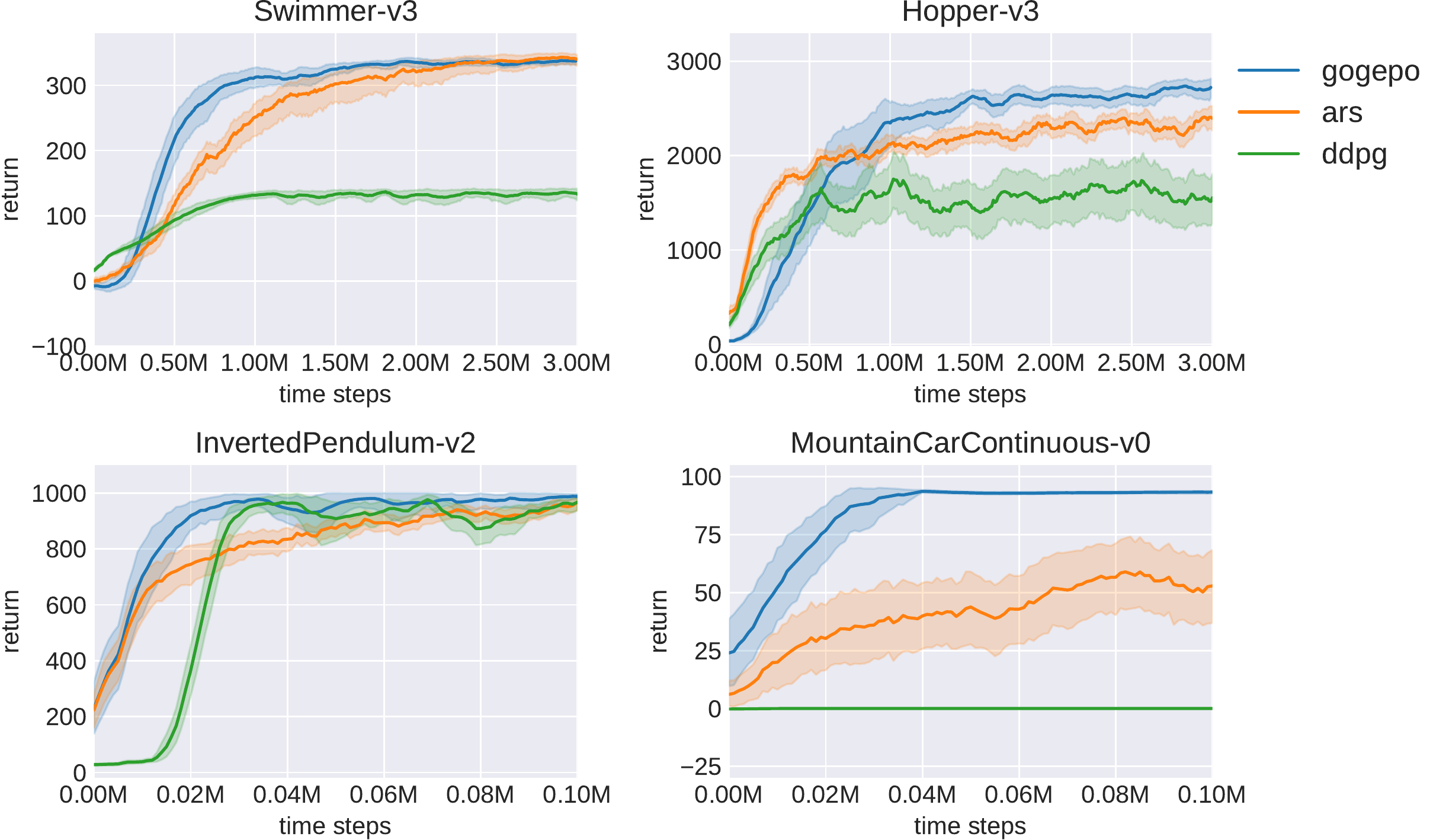} \\
\end{center}
\caption{Performance of policies created with {\approach} (our method), ARS and DDPG over the course of training. Curves show the mean return and 95\% bootstrapped confidence intervals from 20 runs as a function of total environment interactions.}
\label{fig:results}
\end{figure}

We evaluate our method on continuous control tasks from the MuJoCo~\citep{todorov} suite.
Parameter-based Augmented Random Search (ARS)~\citep{mania2018simple} serves as a strong baseline.
We also compare our method to the Deep Deterministic Policy Gradient (DDPG) algorithm~\citep{Silver2014}, another popular method for continuous control tasks.
In the experiments, all policies are MLPs with two hidden layers, each having 256 neurons. 
Our method uses the same set of hyperparameters in all environments.
For ARS, we tune step size, population size, and noise independently for each environment. 
For DDPG, we use the established set of default hyperparameters. 
Details can be found in Appendix~\ref{ap:impl}.

We find that while always asking to generate a policy with return equal to the best return ever seen, there is a slight advantage when asking for more than that.
In particular, we demonstrate that a simple strategy such as ``produce a policy whose return is 20 above the one of the best policy seen so far'' can be very effective.
We present an ablation showing that this strategy is slightly better than the strategy ``produce a policy whose return equal to the one of the best policy seen so far'' in Appendix~\ref{ap:command}.
This suggests that our method's success is not only due to random exploration in parameter space but also to generalization over commands: {\em it learns to understand and exploit the nature of performance improvements in a given environment.}

For our method and ARS, we use observation normalization (see \cite{mania2018simple, faccio2020parameter}).
Furthermore, following ARS, the survival bonus of +1 for every timestep is removed for the Hopper-v3 environment, since for parameter-based methods it leads to the local optimum of staying alive without any movement.

In tasks without fixed episode length, quickly failing bad policies from the early stages of training tend to dominate the replay buffer.
To counteract this, we introduce a recency bias when sampling training batches from the buffer, assigning higher probability to newer policies.
It is treated as an additional hyperparameter. Below we provide an ablation showing the importance of this component. Figure~\ref{fig:results} shows our main experimental result.

Our Algorithm~\ref{alg:pvf} performs competitively in the tested environments.
In Swimmer and Hopper environments, our method learns faster than ARS, while eventually reaching the same asymptotic performance.
In MountainCarContinuous, DDPG is unable to explore the action space, and parameter-based methods quickly learn the optimal policy. 
For a comparison to UDRL with episodic resets, see Appendix~\ref{ap:udrl}.

\begin{figure}[H]

\begin{subfigure}[c]{1.0\textwidth}
  \centering
  \includegraphics[width=1.0\linewidth]{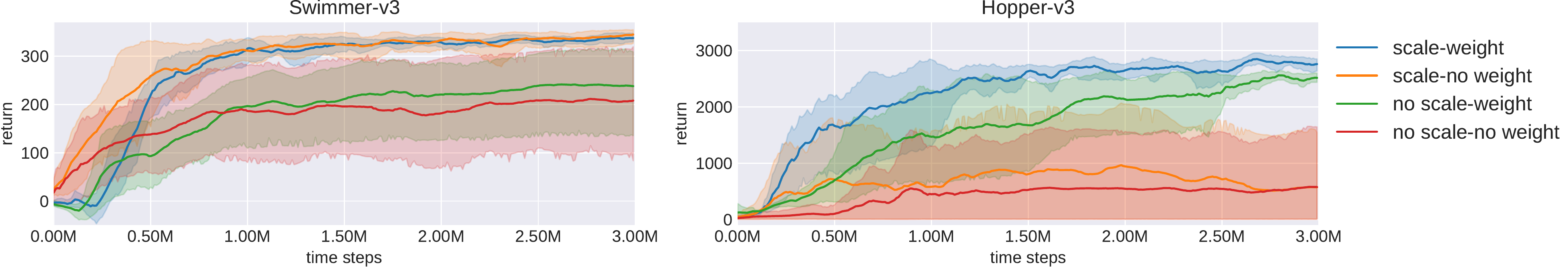}
  \vspace{0.3cm}
    \centering
\end{subfigure}%
\hfill

\caption{Comparison between our algorithm with/without weighted sampling from the replay buffer and output scaling. ``No weight'' denotes uniform sampling from the replay buffer. Average over 5 independent runs and $95\%$ bootstrapped confidence intervals.}
\label{fig:ablation_weighting}
\end{figure}

\paragraph{Obtaining suitable policies from the start}
Randomly initialized policy generators produce weights   far from those of typical initialization schemes. In particular, the standard PyTorch \citep{paszke2019pytorch} initialization is uniform in $[-1/\sqrt{n}, 1/\sqrt{n}]$, where $n$ is the number of neurons in the previous layer, resulting in a distribution uniform in $[-0.0625, 0.0625]$ in the second and last layers. Our network tends to generate much larger weights, roughly uniform in every NN layer. We therefore scale our output such that it is close to the default initialization. Concretely, we multiply for each layer the output of the generator by $2/\sqrt{n}$, where $n$ is the number of neurons in the previous layer. Here we provide an ablation showing that this choice is crucial. Figure~\ref{fig:ablation_weighting} shows the importance of scaling the output of the generator in Swimmer and Hopper. We compare this with and without weighted sampling from the replay buffer. We observe that in Swimmer, output scaling is very important, while in Hopper, most of the performance gain is due to weighted sampling. This choice of output scaling is rather heuristic and does not match the standard PyTorch initialization for all environments. It might happen that a randomly initialized generator produces policies that are difficult to perturb. This exploration issue seems to cause some difficulties for InvertedDoublePendulum, highlighting a possible limitation of our method.

\subsection{Analyzing the generator's learning process}
\label{sec:identity}
\begin{figure}[t]
\begin{center}
\centering
\includegraphics[width=1.\linewidth]{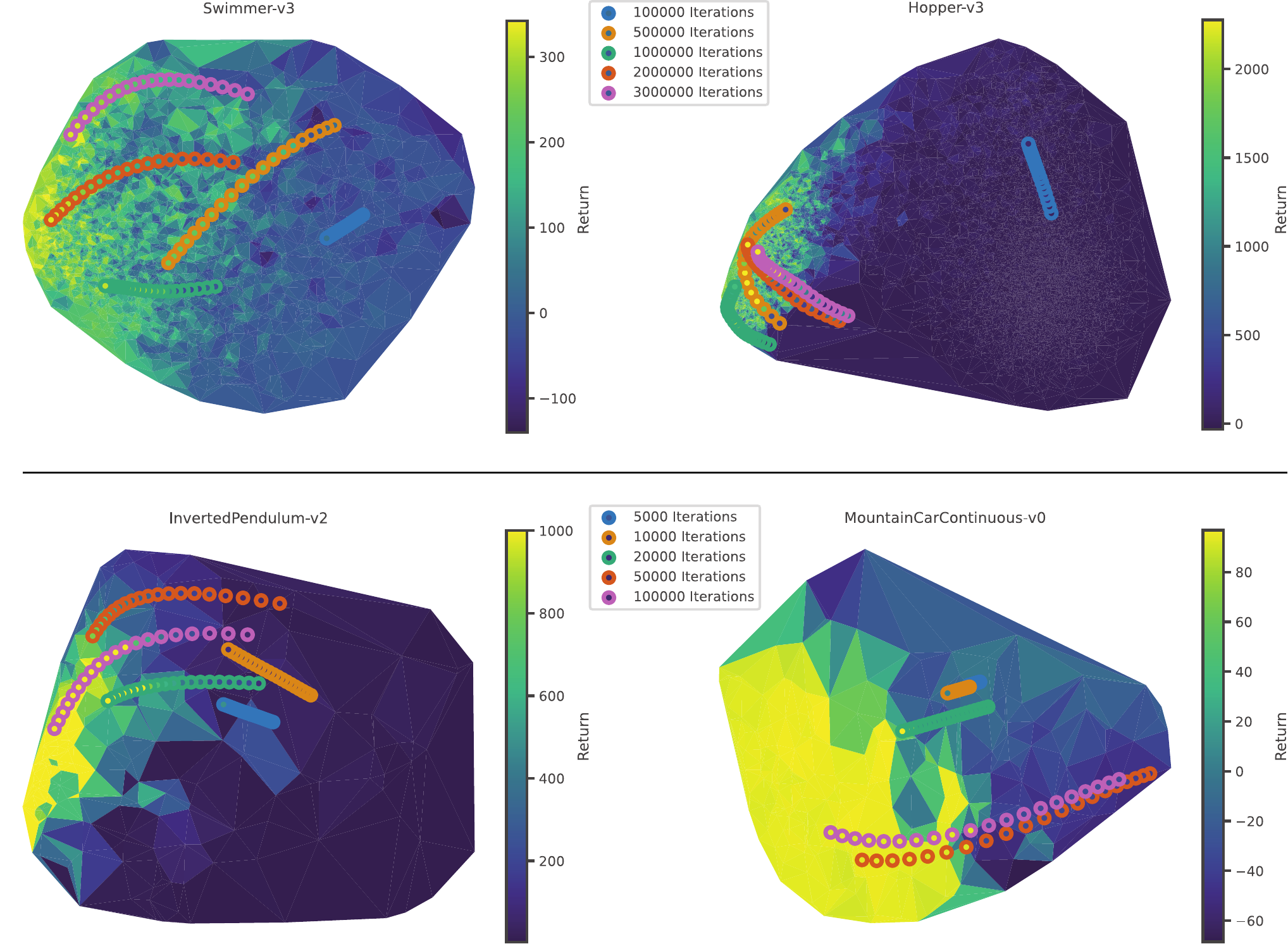}
\end{center}
\caption{Policies generated by the generator during different stages of training. 
The generator is able to produce policies across the whole performance spectrum.
To visualize this, each generator is given 20 return commands ranging from the minimum to the maximum possible return in the environment. 
The color shows the achieved return of each policy. The positions are determined by the probing actions (obtained by the final critic $V_{\textbf{w}}$). 
The background shows the policies in the buffer, i.e., policies observed during training, and total reward. 
Probing actions are reduced to two dimensions by applying PCA to the buffer policies.
}
\label{fig:trajectory_pca}
\end{figure}

\begin{figure}[ht]
     \centering
     \begin{subfigure}[b]{0.4\linewidth}
         \centering
         \includegraphics[width=\linewidth]{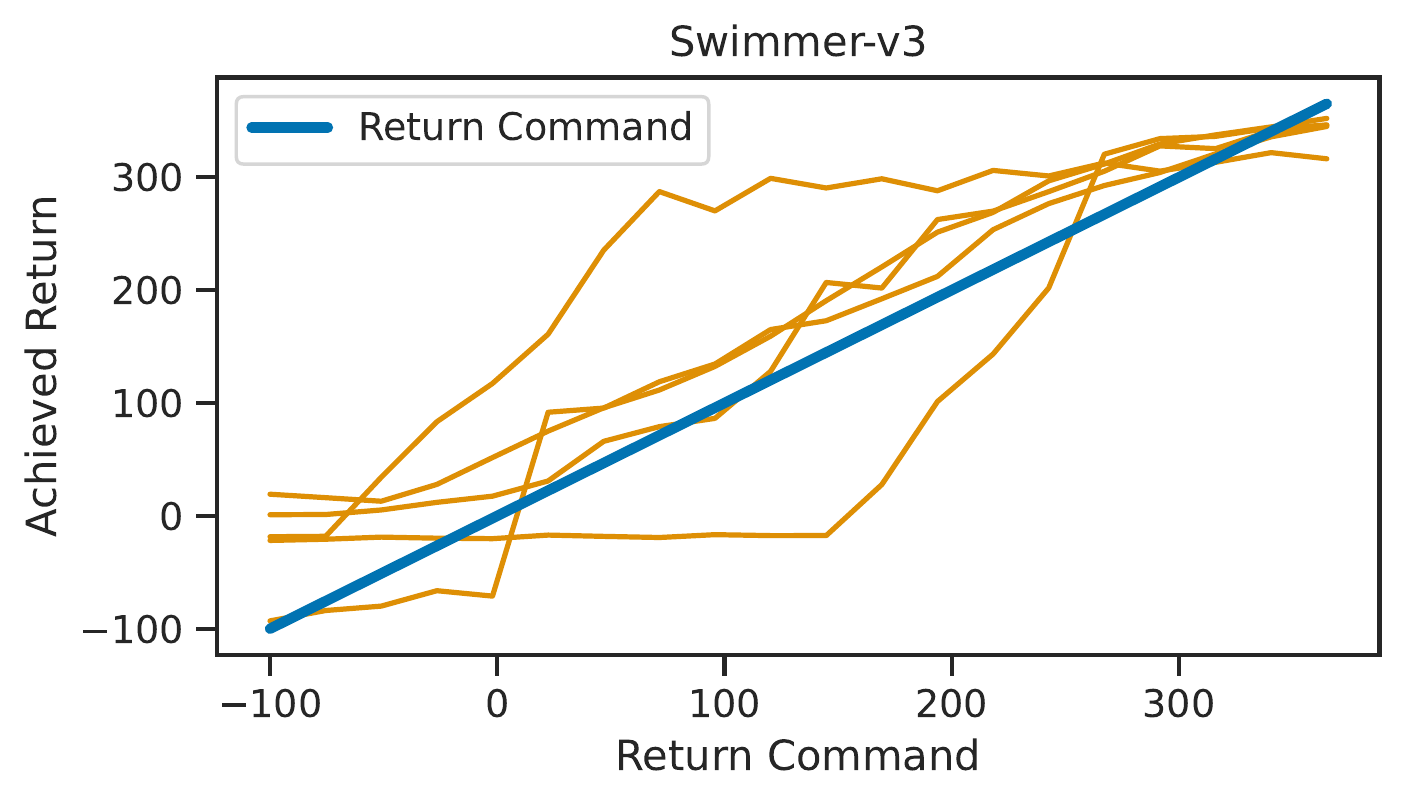}
         \caption{}
         \label{fig:swimmer_identity}
     \end{subfigure}
     \hspace{0.1\linewidth}
     \begin{subfigure}[b]{0.4\linewidth}
         \centering
         \includegraphics[width=\linewidth]{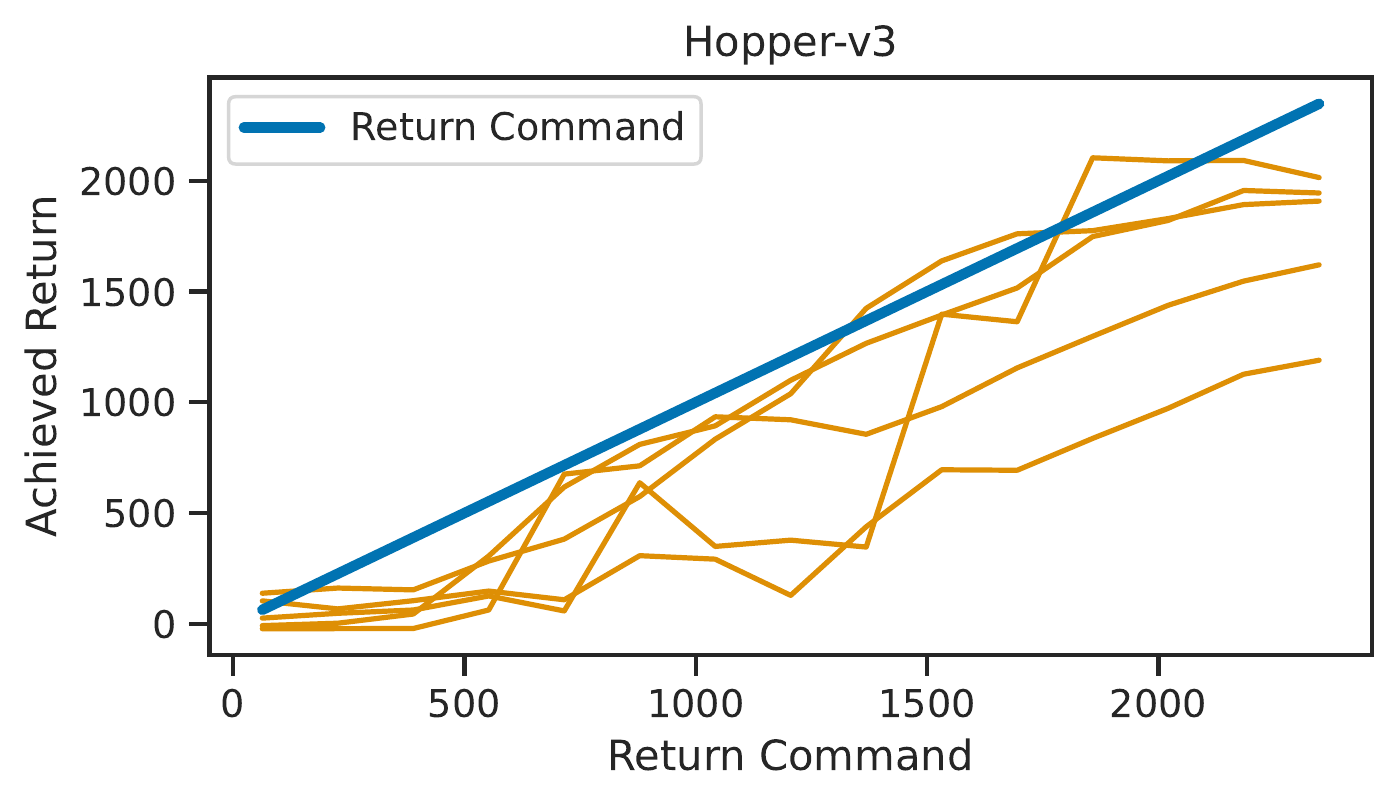}
         \caption{}
         \label{fig:hopper_identity}
     \end{subfigure}
     \begin{subfigure}[b]{0.4\linewidth}
         \centering
         \includegraphics[width=\linewidth]{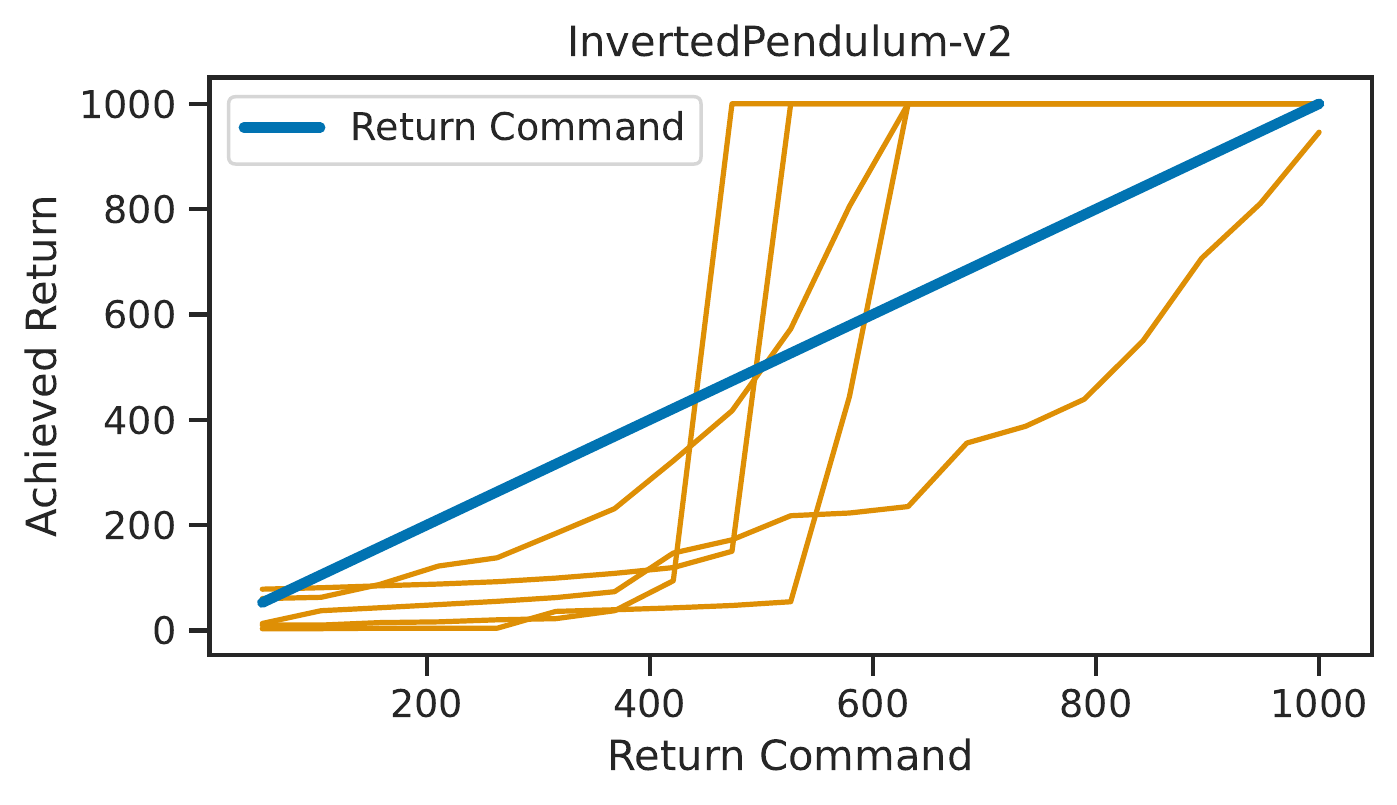}
         \caption{}
         \label{fig:pendulum_identity}
     \end{subfigure}
     \hspace{0.1\linewidth}
     \begin{subfigure}[b]{0.4\linewidth}
         \centering
         \includegraphics[width=\linewidth]{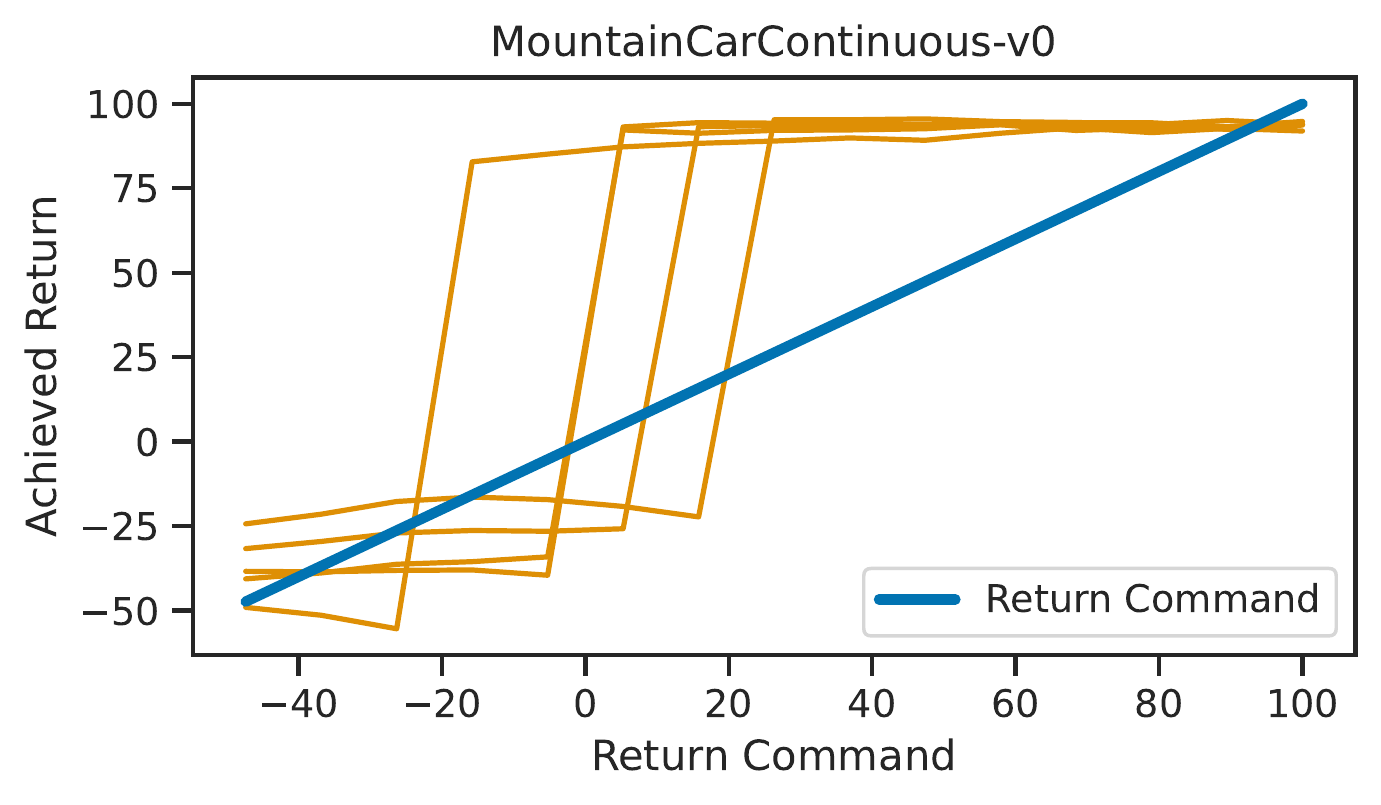}
         \caption{}
         \label{fig:mountain_identity}
     \end{subfigure}
     \caption{
     Achieved returns (mean of 10 episodes) of policies created by fully trained generators as a function of the given return command.
     A perfect generator would produce policies that lie on the diagonal identity line (if the environment permits such returns).
     For each environment, results of five independent runs are shown.
     }
     \label{fig:identity_plots}
\end{figure}

The probing actions created by the fingerprinting mechanism of the value function $V_{\textbf{w}}$ can be seen as a compact meaningful policy embedding useful to visualize policies for a specific environment.
In Figure~\ref{fig:trajectory_pca} we apply PCA to probing actions to show all policies in the buffer after training, as well as policies created by the generator at different stages of training when given the same range of return commands. Policies are colored in line with achieved return. The generator's objective can be seen as finding a trajectory through policy space, defined by the return commands, connecting the lowest with the highest return. In Figure~\ref{fig:trajectory_pca}, this corresponds to a trajectory going from a dark to a bright area.
Indeed, we observe that the generator starts out being confined to the dark region (producing only bad policies) and over the course of training finds a trajectory leading from the darkest (low return) to the brightest (high return) regions. Figure~\ref{fig:identity_plots} shows the the returns achieved by policies that are created by a fully trained generator when given a range of return commands. 
This highlights a feature of the policy generator: while most RL algorithms generate only the best-performing policy, our generator is in principle able to produce by command policies across the whole performance spectrum.
For the environments Swimmer and Hopper (Figures~\ref{fig:swimmer_identity} and \ref{fig:hopper_identity}), this works in a relatively reliable fashion. In Hopper the return used does not include survival bonus. A return of 2000 without survival bonus corresponds roughly to a return of 3000 with survival bonus.

It is worth noting, however, that in some environments it is hard or even impossible to achieve every given intermediate return.
This might be the case, for example, if the optimal policy is much simpler than a slightly sub-optimal one, or if a large reward is given once a goal state is reached.
We can observe this effect for the environments InvertedPendulum and MountainCar---see Figures~\ref{fig:pendulum_identity} and \ref{fig:mountain_identity}.
There the generator struggles to produce the desired identity of return command and achieved return---instead we get something closer to a step function.
However, this does not prevent our method from quickly finding optimal policies in these environments. More details in Appendix~\ref{ap:visual}.

\section{Related Work}
\label{sec:related_work}
\paragraph{Policy conditioned value functions}
Compared to standard value functions conditioned on a specific policy, policy-conditioned value functions generate values across several policies~\citep{faccio2020parameter,harb2020policy}.
This has been used to directly maximize the value using gradient ascent in the policy parameters.
Here we use it to evaluate any policy generated by our policy generator.
In contrast to previous work, this allows for generating policies of arbitrary quality in a zero-shot manner, without any gradient-based iterative training procedure.

\paragraph{Hindsight and Upside Down RL}
Upside Down RL (UDRL) transforms the RL problem into a supervised learning problem by conditioning the policy on commands such as ``achieve a desired return''~\citep{schmidhuber2019reinforcement,srivastava2019training}.
The required dataset of states, actions, and rewards can be collected online during iterative improvements of the policy~\citep{srivastava2019training}, or offline~\citep{janner2021reinforcement,chen2021decision}.
UDRL methods are related to hindsight RL where the  commands correspond to desired goal states in the environment~\citep{schmidhuber1991learning,kaelbling1993learning,andrychowicz2017hindsight,rauber2018hindsight}.
Instead of optimizing the policy to achieve a desired reward in action space, our method {\approach} evaluates the generated policies in command space.
This is done by generating, conditioning on a command, a policy that is then evaluated using a parameter-based value function and trained to match the command to the evaluated return. This side-steps the issue with multi-modality in certain types of UDRL for episodic environments, where a command may be achieved through many different behaviors, and fitting the policy to varying actions may lead to sub-optimal policies.

\paragraph{Fast Weight Programmers and HyperNetworks}
The idea of using a neural network (NN) to generate weight changes for another NN dates back to Fast Weight Programmers (FWPs)~\citep{schmidhuber1992learning,schmidhuber1993self}, later scaled up to deeper neural networks under the name of hypernetworks~\citep{Ha2016}.
While in traditional NNs the weight matrix remains fixed after training, FWPs make these weights context-dependent. 
More generally, FWPs can be used as neural functions that involve multiplicative interactions and parameter sharing~\citep{kirsch2020meta}.
When updated in recurrent fashion, FWPs can be used as memory mechanisms.
Linear transformers are a type of FWP where information is stored through outer products of keys and values~\citep{schlag2021linear,schmidhuber1992learning}.
FWPs are used in the context of memory-based meta learning~\citep{schmidhuber1993self,miconi2018differentiable,Gregor2020,kirsch2020meta,irie2021a,kirsch2021introducing}, predicting parameters for varying architectures~\citep{knyazev2021parameter}, and reinforcement learning~\citep{Gomez:2005:CRN:1068009.1068092,najarro2020meta,kirsch2021introducing}.
In contrast to all of these approaches, ours uses FWPs to generate policies conditioning on a command (target return).

\section{Conclusion and Future Work}
\label{sec:conclusion}
Our {\approach} is an RL framework for generating  policies yielding given desired returns. 
Hypernetworks in conjunction with fingerprinting-based value functions can be used to train a Fast Weight Programmer through supervised learning to directly generate parameters of a policy that achieves a given return.
By iteratively asking for higher returns than those observed so far, our algorithm trains the generator to produce highly performant policies from scratch. 
Empirically, {\approach} is competitive with ARS and DDPG on continuous control tasks, and able to generate policies with any desired return. Future work will consider context commands other than those asking for particular returns, as well as generators based on
latent variable models (e.g., conditional variational autoencoders) allowing for capturing diverse sets of policies, to improve exploration of complex RL environments.

\begin{ack}
We thank Kazuki Irie, Mirek Strupl, Dylan Ashley, Róbert Csordás, Aleksandar Stanić and Anand Gopalakrishnan for their feedback. This work was supported by the ERC Advanced Grant (no: 742870) and by the Swiss National Supercomputing Centre (CSCS, projects: s1090, s1154). We also thank NVIDIA Corporation for donating a DGX-1 as part of the Pioneers of AI Research Award and to IBM for donating a Minsky machine.
\end{ack}

\typeout{}  
\bibliography{main}

\begin{thebibliography}{}

\bibitem[Achiam, 2018]{SpinningUp2018}
Achiam, J. (2018).
\newblock {Spinning Up in Deep Reinforcement Learning}.

\bibitem[Andrychowicz et~al., 2017]{andrychowicz2017hindsight}
Andrychowicz, M., Wolski, F., Ray, A., Schneider, J., Fong, R., Welinder, P.,
  McGrew, B., Tobin, J., Pieter~Abbeel, O., and Zaremba, W. (2017).
\newblock {Hindsight Experience Replay}.
\newblock In {\em NeurIPS}.

\bibitem[Chen et~al., 2021]{chen2021decision}
Chen, L., Lu, K., Rajeswaran, A., Lee, K., Grover, A., Laskin, M., Abbeel, P.,
  Srinivas, A., and Mordatch, I. (2021).
\newblock Decision transformer: Reinforcement learning via sequence modeling.
\newblock {\em Advances in neural information processing systems}, 34.

\bibitem[Faccio et~al., 2020]{faccio2020parameter}
Faccio, F., Kirsch, L., and Schmidhuber, J. (2020).
\newblock Parameter-based value functions.
\newblock {\em arXiv preprint arXiv:2006.09226}.

\bibitem[Ghosh et~al., 2019]{ghosh2020learning}
Ghosh, D., Gupta, A., Reddy, A., Fu, J., Devin, C., Eysenbach, B., and Levine,
  S. (2019).
\newblock Learning to reach goals via iterated supervised learning.

\bibitem[Gomez and Schmidhuber, 2005]{Gomez:2005:CRN:1068009.1068092}
Gomez, F.~J. and Schmidhuber, J. (2005).
\newblock Co-evolving recurrent neurons learn deep memory pomdps.
\newblock In {\em Proceedings of the 7th Annual Conference on Genetic and
  Evolutionary Computation}, GECCO '05, pages 491--498, New York, NY, USA. ACM.

\bibitem[Gregor, 2020]{Gregor2020}
Gregor, K. (2020).
\newblock {Finding online neural update rules by learning to remember}.
\newblock {\em arXiv preprint arXiv:2003.03124}.

\bibitem[Ha et~al., 2016]{Ha2016}
Ha, D., Dai, A., and Le, Q.~V. (2016).
\newblock {HyperNetworks}.
\newblock In {\em International Conference on Learning Representations}.

\bibitem[Harb et~al., 2020]{harb2020policy}
Harb, J., Schaul, T., Precup, D., and Bacon, P.-L. (2020).
\newblock Policy evaluation networks.
\newblock {\em arXiv preprint arXiv:2002.11833}.

\bibitem[Hesterberg, 1988]{hesterberg1988advances}
Hesterberg, T.~C. (1988).
\newblock {\em Advances in importance sampling}.
\newblock PhD thesis, Stanford University.

\bibitem[Irie et~al., 2021]{irie2021a}
Irie, K., Schlag, I., Csord{\'a}s, R., and Schmidhuber, J. (2021).
\newblock A modern self-referential weight matrix that learns to modify itself.
\newblock In {\em Deep RL Workshop NeurIPS 2021}.

\bibitem[Janner et~al., 2021]{janner2021reinforcement}
Janner, M., Li, Q., and Levine, S. (2021).
\newblock {Offline Reinforcement Learning as One Big Sequence Modeling
  Problem}.
\newblock In {\em NeurIPS}.

\bibitem[Kaelbling, 1993]{kaelbling1993learning}
Kaelbling, L.~P. (1993).
\newblock Learning to achieve goals.
\newblock In {\em IJCAI}.

\bibitem[Kaelbling et~al., 1996]{Kaelbling:96}
Kaelbling, L.~P., Littman, M.~L., and Moore, A.~W. (1996).
\newblock Reinforcement learning: A survey.
\newblock {\em Journal of artificial intelligence research}, 4:237--285.

\bibitem[Kirsch et~al., 2022]{kirsch2021introducing}
Kirsch, L., Flennerhag, S., van Hasselt, H., Friesen, A., Oh, J., and Chen, Y.
  (2022).
\newblock {Introducing Symmetries to Black Box Meta Reinforcement Learning}.
\newblock In {\em Proceedings of the AAAI Conference on Artificial
  Intelligence}.

\bibitem[Kirsch and Schmidhuber, 2021]{kirsch2020meta}
Kirsch, L. and Schmidhuber, J. (2021).
\newblock Meta learning backpropagation and improving it.
\newblock {\em Advances in Neural Information Processing Systems}, 34.

\bibitem[Knyazev et~al., 2021]{knyazev2021parameter}
Knyazev, B., Drozdzal, M., Taylor, G.~W., and Romero~Soriano, A. (2021).
\newblock Parameter prediction for unseen deep architectures.
\newblock {\em Advances in Neural Information Processing Systems}, 34.

\bibitem[Kůrková and Kainen, 1994]{kuurkova1994functionally}
Kůrková, V. and Kainen, P.~C. (1994).
\newblock Functionally equivalent feedforward neural networks.
\newblock {\em Neural Computation}, 6(3):543--558.

\bibitem[Mania et~al., 2018]{mania2018simple}
Mania, H., Guy, A., and Recht, B. (2018).
\newblock Simple random search of static linear policies is competitive for
  reinforcement learning.
\newblock In {\em Advances in Neural Information Processing Systems}, pages
  1800--1809.

\bibitem[Miconi et~al., 2018]{miconi2018differentiable}
Miconi, T., Stanley, K., and Clune, J. (2018).
\newblock Differentiable plasticity: training plastic neural networks with
  backpropagation.
\newblock In {\em International Conference on Machine Learning}, pages
  3559--3568. PMLR.

\bibitem[Najarro and Risi, 2020]{najarro2020meta}
Najarro, E. and Risi, S. (2020).
\newblock Meta-learning through hebbian plasticity in random networks.
\newblock {\em Advances in Neural Information Processing Systems},
  33:20719--20731.

\bibitem[Papini et~al., 2019]{papini2019optimistic}
Papini, M., Metelli, A.~M., Lupo, L., and Restelli, M. (2019).
\newblock Optimistic policy optimization via multiple importance sampling.
\newblock In {\em 36th International Conference on Machine Learning},
  volume~97, pages 4989--4999.

\bibitem[Paszke et~al., 2019]{paszke2019pytorch}
Paszke, A., Gross, S., Massa, F., Lerer, A., Bradbury, J., Chanan, G., Killeen,
  T., Lin, Z., Gimelshein, N., Antiga, L., et~al. (2019).
\newblock Pytorch: An imperative style, high-performance deep learning library.
\newblock {\em Advances in neural information processing systems}, 32.

\bibitem[Puterman, 2014]{puterman2014markov}
Puterman, M.~L. (2014).
\newblock {\em Markov decision processes: discrete stochastic dynamic
  programming}.
\newblock John Wiley \& Sons.

\bibitem[Rauber et~al., 2018]{rauber2018hindsight}
Rauber, P., Ummadisingu, A., Mutz, F., and Schmidhuber, J. (2018).
\newblock Hindsight policy gradients.
\newblock In {\em International Conference on Learning Representations}.

\bibitem[Salimans et~al., 2017]{salimans2017evolution}
Salimans, T., Ho, J., Chen, X., Sidor, S., and Sutskever, I. (2017).
\newblock Evolution strategies as a scalable alternative to reinforcement
  learning.
\newblock {\em arXiv preprint arXiv:1703.03864}.

\bibitem[Schaul et~al., 2015]{Schaul:2015:UVF:3045118.3045258}
Schaul, T., Horgan, D., Gregor, K., and Silver, D. (2015).
\newblock Universal value function approximators.
\newblock In {\em Proceedings of the 32Nd International Conference on
  International Conference on Machine Learning - Volume 37}, ICML'15, pages
  1312--1320. JMLR.org.

\bibitem[Schlag et~al., 2021]{schlag2021linear}
Schlag, I., Irie, K., and Schmidhuber, J. (2021).
\newblock Linear transformers are secretly fast weight programmers.
\newblock In {\em International Conference on Machine Learning}, pages
  9355--9366. PMLR.

\bibitem[Schmidhuber, 1990]{Schmidhuber:90sandiego}
Schmidhuber, J. (1990).
\newblock An on-line algorithm for dynamic reinforcement learning and planning
  in reactive environments.
\newblock In {\em Proc. IEEE/INNS International Joint Conference on Neural
  Networks, San Diego}, volume~2, pages 253--258.

\bibitem[Schmidhuber, 1991]{schmidhuber1991learning}
Schmidhuber, J. (1991).
\newblock Learning to generate sub-goals for action sequences.
\newblock In {\em Artificial neural networks}, pages 967--972.

\bibitem[Schmidhuber, 1992]{schmidhuber1992learning}
Schmidhuber, J. (1992).
\newblock Learning to control fast-weight memories: An alternative to dynamic
  recurrent networks.
\newblock {\em Neural Computation}, 4(1):131--139.

\bibitem[Schmidhuber, 1993]{schmidhuber1993self}
Schmidhuber, J. (1993).
\newblock A ‘self-referential’weight matrix.
\newblock In {\em International Conference on Artificial Neural Networks},
  pages 446--450. Springer.

\bibitem[Schmidhuber, 2015]{learningtothink2015}
Schmidhuber, J. (2015).
\newblock On learning to think: Algorithmic information theory for novel
  combinations of reinforcement learning controllers and recurrent neural world
  models.
\newblock {\em Preprint arXiv:1511.09249}.

\bibitem[Schmidhuber, 2019]{schmidhuber2019reinforcement}
Schmidhuber, J. (2019).
\newblock {Reinforcement Learning Upside Down: Don't Predict Rewards--Just Map
  Them to Actions}.
\newblock {\em arXiv:1912.02875}.

\bibitem[Schmidhuber and Huber, 1991]{SchmidhuberHuber:91}
Schmidhuber, J. and Huber, R. (1991).
\newblock Learning to generate artificial fovea trajectories for target
  detection.
\newblock {\em International Journal of Neural Systems}, 2(1 \& 2):135--141.
\newblock (Based on TR FKI-128-90, TUM, 1990).

\bibitem[Sehnke et~al., 2008]{sehnkepgpecontrol}
Sehnke, F., Osendorfer, C., R{\"u}ckstie{\ss}, T., Graves, A., Peters, J., and
  Schmidhuber, J. (2008).
\newblock Policy gradients with parameter-based exploration for control.
\newblock In K{\r{u}}rkov{\'a}, V., Neruda, R., and Koutn{\'i}k, J., editors,
  {\em Artificial Neural Networks - ICANN 2008}, pages 387--396, Berlin,
  Heidelberg. Springer Berlin Heidelberg.

\bibitem[Sehnke et~al., 2010]{sehnke_parameterexploring_2010}
Sehnke, F., Osendorfer, C., Rückstieß, T., Graves, A., Peters, J., and
  Schmidhuber, J. (2010).
\newblock Parameter-exploring policy gradients.
\newblock {\em Neural Networks}, 23(4):551--559.

\bibitem[Silver et~al., 2014]{Silver2014}
Silver, D., Lever, G., Heess, N., Degris, T., Wierstra, D., and Riedmiller, M.
  (2014).
\newblock Deterministic policy gradient algorithms.
\newblock In {\em Proceedings of the 31st International Conference on
  International Conference on Machine Learning - Volume 32}, ICML'14, pages
  I--387--I--395. JMLR.org.

\bibitem[Srivastava et~al., 2019]{srivastava2019training}
Srivastava, R.~K., Shyam, P., Mutz, F., Ja{\'s}kowski, W., and Schmidhuber, J.
  (2019).
\newblock {Training Agents Using Upside-down Reinforcement Learning}.
\newblock In {\em NeurIPS Deep RL Workshop}.

\bibitem[Stratonovich, 1960]{stratonovich1960}
Stratonovich, R. (1960).
\newblock Conditional {Markov} processes.
\newblock {\em Theory of Probability And Its Applications}, 5(2):156--178.

\bibitem[{\v{S}}trupl et~al., 2022]{vstrupl2022upside}
{\v{S}}trupl, M., Faccio, F., Ashley, D.~R., Schmidhuber, J., and Srivastava,
  R.~K. (2022).
\newblock Upside-down reinforcement learning can diverge in stochastic
  environments with episodic resets.
\newblock {\em arXiv preprint arXiv:2205.06595}.

\bibitem[Sutton and Barto, 2018]{Sutton:2018:RLI:3312046}
Sutton, R.~S. and Barto, A.~G. (2018).
\newblock {\em Reinforcement Learning: An Introduction}.
\newblock A Bradford Book, USA.

\bibitem[Sutton et~al., 1999]{Sutton1999}
Sutton, R.~S., McAllester, D., Singh, S., and Mansour, Y. (1999).
\newblock Policy gradient methods for reinforcement learning with function
  approximation.
\newblock In {\em Proceedings of the 12th International Conference on Neural
  Information Processing Systems}, NIPS'99, pages 1057--1063, Cambridge, MA,
  USA. MIT Press.

\bibitem[Todorov et~al., 2012]{todorov}
Todorov, E., Erez, T., and Tassa, Y. (2012).
\newblock Mujoco: A physics engine for model-based control.
\newblock In {\em 2012 IEEE/RSJ International Conference on Intelligent Robots
  and Systems}, pages 5026--5033.

\bibitem[van Hasselt, 2012]{hasselt2012}
van Hasselt, H. (2012).
\newblock Reinforcement learning in continuous state and action spaces.
\newblock In Wiering, M. and van Otterlo, M., editors, {\em Reinforcement
  Learning}, pages 207--251. Springer.

\bibitem[Veach and Guibas, 1995]{veach1995optimally}
Veach, E. and Guibas, L.~J. (1995).
\newblock Optimally combining sampling techniques for monte carlo rendering.
\newblock In {\em Proceedings of the 22nd annual conference on Computer
  graphics and interactive techniques}, pages 419--428.

\bibitem[von Oswald et~al., 2020]{von2020continual}
von Oswald, J., Henning, C., Sacramento, J., and Grewe, B.~F. (2020).
\newblock Continual learning with hypernetworks.
\newblock In {\em 8th International Conference on Learning Representations
  (ICLR 2020)(virtual)}. International Conference on Learning Representations.

\bibitem[Williams, 1992]{williams1992simple}
Williams, R.~J. (1992).
\newblock Simple statistical gradient-following algorithms for connectionist
  reinforcement learning.
\newblock In {\em Reinforcement Learning}, pages 5--32. Springer.

\end{thebibliography}
\bibliographystyle{apalike}


\clearpage
\appendix
\section{Implementation details}
\label{ap:impl}

\subsection{Hyperparameters}
\label{ap:hyperparams}
Here we report the hyperparameters used for {\approach} and the baselines. 
For DDPG, we use the spinning-up RL implementation ~\citep{SpinningUp2018}, whose results are on par with the best reported results. 
For ARS, we use the implementation of the authors~\citep{mania2018simple}, adapted to Deep NN policies.

\paragraph{Shared hyperparameters} The table below shows hyperparameters relevant to at least two of the three methods. 
They stay fixed across environments.

\begin{center}
\small
{\renewcommand{\arraystretch}{1.5}
\begin{tabular}{| p{0.33\textwidth} || p{0.16\textwidth} | p{0.20\textwidth} | p{0.17\textwidth} |} 
 \hline
 Hyperparameter & ARS & {\approach} & DDPG \\  
 \hline\hline
 Policy Architecture & \multicolumn{3}{c|}{MLP, 2 hidden layers, 256 neurons each, with bias} \\ 
 \hline
 Policy Nonlinearity & \multicolumn{2}{c|}{tanh} & ReLU \\
 \hline
 Value Function Architecture & & \multicolumn{2}{p{0.4\textwidth}|}{MLP, 2 hidden layers, 256 neurons each, with bias} \\
 \hline
 Value Function Nonlinearity & & \multicolumn{2}{c|}{ReLU} \\
 \hline
 Initialization MLPs & & PyTorch default (for value function) & PyTorch default (for actor \& critic) \\  
 \hline
 Batch Size & & 16 & 128 \\
  \hline
 Optimizer & & \multicolumn{2}{c|}{Adam}\\
 \hline
 Learning Rate Actor/Generator & &2e-6  & 1e-3 \\
 \hline
 Learning Rate Value Function & & 5e-3 & 1e-3 \\
 \hline
 Exploration Noise Scale & tuned (see below) & 0.1 in parameter space & 0.1 in action space \\
 \hline
 Update Frequency Actor/Generator & every batch & every episode  & every 50 time steps \\
 \hline
 Update Frequency Value Function & & every episode & every 50 time steps \\
 \hline
  Number of Actor/Generator Updates & & 20 & 50 \\
 \hline
 Number of Value Function Updates & & 5 & 50 \\
 \hline
 Replay Buffer Size & & 10k & 100k \\
 \hline
 Discount Factor & \multicolumn{2}{c|}{1}& 0.99 \\
 \hline
 Survival Reward Adjustment &  \multicolumn{2}{c|}{True (for Hopper)} & False \\
  \hline
  Observation Normalization &  \multicolumn{2}{c|}{True} & False \\
 \hline
 Environmental interactions &  \multicolumn{3}{p{0.6\textwidth}|}{100k for InvertedPendulum and MountCarContinuous, 3M for all other environments} \\
 \hline
\end{tabular}}
\end{center}

\paragraph{Hyperparameters for specific algorithms}
Fixed across environments:

{\approach}:
\begin{itemize}
    \item Architecture of the networks $H$ in the generator: MLP with bias, two hidden layers of size 256, ReLU nonlinearity, no output activation function
    \item Size of learnable hypernetwork embeddings $z^j_{mn}$: $8$
    \item Size of slices $s^j_{mn}$ produced by the hypernetwork: $16 \times 16$
    \item Number of probing states: 200
    \item Initialization of probing states: Uniformly random in $[0,1)$
    \item Priority sampling from replay buffer: True, with weights $1/x^{1.1}$, where x is the number of episodes since the data was stored in the buffer
\end{itemize}

DDPG:
\begin{itemize}
    \item Start-steps (random actions): 10000 time steps
    \item Update after (no training): 1000 time steps
    \item Polyak parameter: 0.995
\end{itemize}

\paragraph{Tuned hyperparameters} For ARS, we tune the following hyperparameters for each environment separately using grid search: 
\begin{itemize}
\item Step size for ARS: tuned with values in $\{ 1e-2, 1e-3, 1e-4 \}$
\item Number of directions and elite directions for ARS: tuned with values in $\{ [1,1], [8,4], [8,8], [32,4], [32,16], [64,8], [64,32]  \}$, where the first element denotes the number of directions and the second element the number of elite directions
\item Noise for exploration in ARS: tuned with values in $\{ 0.1, 0.05, 0.025 \}$
\end{itemize}
Here we report the best hyperparameters found for each environment:
\begin{center}
\small
{\renewcommand{\arraystretch}{1.5}
\begin{tabular}{| p{0.29\textwidth} || p{0.13\textwidth} | p{0.13\textwidth} | p{0.13\textwidth} | p{0.14\textwidth}|} 
 \hline
 ARS Hyperparameter & Swimmer & Hopper & Inverted-Pendulum & MountainCar-Continuous \\  
 \hline\hline
 Step Size & 0.01 & 0.01 & 0.001 & 0.01 \\ 
 \hline
 Number of Directions, Number of Elite Directions & (8, 4) & (8, 4) & (1, 1) & (1, 1) \\
 \hline
 Exploration Noise Scale & 0.05 & 0.05 & 0.025 & 0.05 \\
 \hline
\end{tabular}}
\end{center}

\paragraph{UDRL}
For UDRL we use a previous implementation ~\citep{srivastava2019training} for discrete control environments, and implemented additional classes to use it in continuous control tasks with episodic resets (although the original UDRL report~\citep{schmidhuber2019reinforcement} focused on continuous control in single-life settings without resets). We use the previous hyperparameters~\citep{srivastava2019training} and tune learning rate (in $\{1e-3, 1e-4, 1e-5\}$), activation (ReLU, tanh), and their ``last\_few'' parameter (1, 10, 100), which is used to select the command for exploration. 
For Swimmer, we are not able to reproduce the performance with the original reported hyperparameters.
Like for the other algorithms, we use an NN with 2 hidden layers and 256 neurons per layer. 
Below we report the best hyperparameters found for UDRL. 

\begin{center}
\small
{\renewcommand{\arraystretch}{1.5}
\begin{tabular}{| p{0.29\textwidth} || p{0.13\textwidth} | p{0.13\textwidth} | p{0.13\textwidth} | p{0.14\textwidth}|} 
 \hline
 UDRL Hyperparameter & Swimmer & Hopper & Inverted-Pendulum & MountainCar-Continuous \\  
 \hline\hline
 Nonlinearity & ReLU & ReLU & tanh & ReLU \\ 
 \hline
 Learning Rate & 1e-3 & 1e-5 & 1e-3 & 1e-5 \\
 \hline
 Last Few & 10 & 10 & 1 & 1 \\
 \hline
\end{tabular}}
\end{center}

\subsection{Generator implementation details}
\label{ap:impl_gen}
\paragraph{Generating bias vectors}
Here we describe how to generate the bias vectors of the policies, which is not explicitly mentioned in section~\ref{sec:hn}.
Analogously to Equations~\ref{eq:hn_slice} and \ref{eq:hn_cat}, the embeddings $z^j_{mn}$ are fed to a dedicated bias-generating network $H_\chi$ that produces slices of the shape $f \times 1$, and those slices are concatenated.
Since we have a two-dimensional grid of learned embeddings $z$ (see Figure~\ref{fig:hypernetwork}), we take the mean across the input dimensions of the concatenated slices so that we end up with a bias vector (and not a matrix).

\subsection{GPU usage / compute}
We use cloud computing resources for our experiments. 
Our nodes have an Intel Xeon 12 core CPU and an NVIDIA Tesla P100 GPU with 16GB of memory.
We were able to run four {\approach} experiments on one node in parallel.
Our estimate of computation time for the main results is 40 node hours.

\section{Experimental details}
\label{ap:exp}

\subsection{Additional experimental results}
\label{ap:udrl}
\begin{figure}[t]
\begin{center}
\centering
\includegraphics[width=0.8\linewidth]{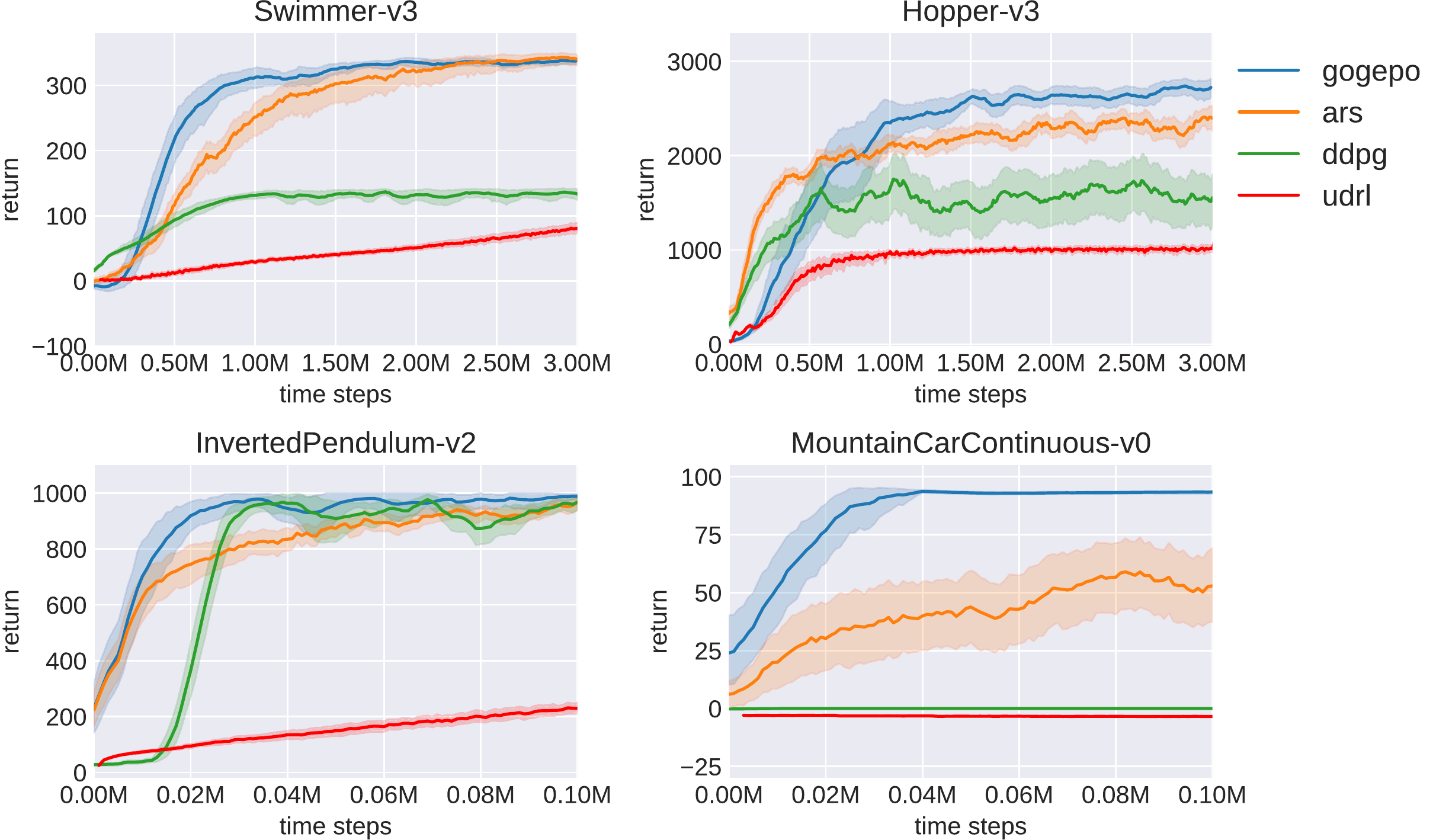} \\
\end{center}
\caption{Performance of policies created with {\approach} (our method), ARS, DDPG and UDRL over the course of training. Curves show the mean return and 95\% bootstrapped confidence intervals from 20 runs as a function of total environmental interactions.}
\label{fig:results_new}
\end{figure}

In Figure~\ref{fig:results_new} we provide additional results to compare our method to UDRL with episodic resets.
We confirm that UDRL is not sample efficient for continuous control in environments with episodic resets~\citep{schmidhuber2019reinforcement}, in line with previous experimental results. We argue that the multimodality issue discussed in the introduction is the main issue with UDRL.

\label{ap:exp_results}

\subsection{Main experiments on MuJoCo}
\label{ap:mujoco}
For ARS and UDRL, the best hyperparameters for each environment are determined by running the algorithm with each hyperparameter configuration across 5 random seeds. 
The best configurations are those reported in section~\ref{ap:hyperparams} 
We use them for the final 20 evaluation runs shown in our main results.
For DDPG and {\approach}, we use the same hyperparameters for all environments.
For 10 episodes, Figures~\ref{fig:results} and \ref{fig:results_new} evaluate each run every 10000 time steps for Swimmer and Hopper, every 1000 steps for InvertedPendulum and MountainCarContinuous. 
Table~\ref{tab:final_ret} shows the final return and standard deviation of each algorithm.

\begin{table}[!h]
  \caption{Final return (average over final 20 evaluations)}
  \begin{tabular}{lrcccc}
    \hline
    \textbf{Environment}  & & {\approach} & ARS & DDPG & UDRL\\
    \hline
    Swimmer-v3 &                &$\bm{334 \pm 16}$       & $\bm{342 \pm 21}$      & $129 \pm 25$      & $78 \pm 17$ \\
    MountainCarContinuous-v0 &  &$\bm{93 \pm 1}$         & $55 \pm 33$       & $-1 \pm 0.01$     & $-3 \pm 0.3$ \\
    Hopper-v3 &                 &$\bm{2589 \pm 300}$     & $\bm{2340 \pm 199}$    & $1634 \pm 1036$    & $1010 \pm 78$ \\
    InvertedPendulum-v2 &       &$\bm{980 \pm 40}$       & $\bm{936 \pm 42}$      & $\bm{960 \pm 175}$     & $219 \pm 299$ \\
    \hline
  \end{tabular}
    \label{tab:final_ret}

\end{table}

\subsection{Details on generated policy visualization}
\label{ap:visual}
To create Figure~\ref{fig:trajectory_pca}, we perform Principal Component Analysis (PCA) on the probing actions of all policies in the buffer after training.
The first two principal components indicate a policy's position in our visualization.
Using Delaunay triangulation, we assign an area to every policy and color it according to its achieved return.
We then take the generator at different stages of training (of the same run).
Each of these generators is given a set of 20 commands, evenly spaced across the range of possible returns ($[-100, 365]$ for Swimmer, $[-100, 3000]$ for Hopper, $[0, 1000]$ for InvertedPendulum and $[-100, 100]$ for MountainCarContinuous).
The resulting policies are plotted using probing actions on the probing states of the fully trained value function $V_\mathbf{w}$ (and the same PCA).

\subsection{Command strategies}
\label{ap:command}
In early experiments, we tried an alternative approach using Importance Sampling~\citep{hesterberg1988advances} estimators.
Given a mixture of weights $\beta_i(\theta)$, we considered estimators of the form $\hat{J}(c', w') = \sum_{i=1}^N \beta_i(\theta_i) \frac{p(\theta_i | c'; w')}{p(\theta_i | c_i; w_i)} r_i$, which provides an unbiased estimate of the performance of a policy produced by a generator with parameters $w'$ and command $c'$, using past data derived from old generators with different commands.
Maximizing $\hat{J}(c', w')$ with respect to the command $c'$ should yield commands encouraging the generator to produce highly performant policies.
We tested this using the Balance Heuristic~\citep{veach1995optimally} estimator for $\beta_k$, which is known to have small variance~\citep{papini2019optimistic}.
However, in our experiments we observed that generators using such command strategies did not significantly outperform the simple strategy mentioned earlier. 

\begin{figure}[H]

\begin{subfigure}[c]{1.0\textwidth}
  \centering
  \includegraphics[width=1.0\linewidth]{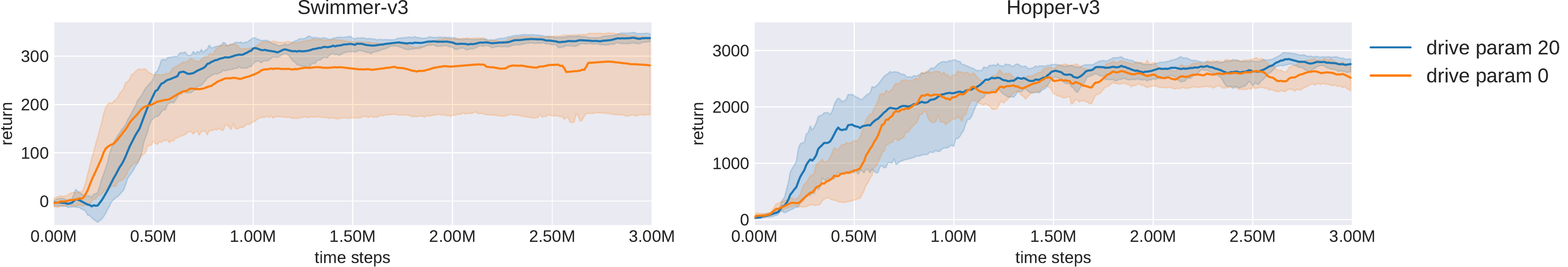}
  \vspace{0.3cm}
    \centering
\end{subfigure}%
\hfill

\caption{Comparison of variants of our algorithm with/without drive parameter for command exploration. Average over 5 independent runs and $95\%$ bootstrapped confidence intervals.}
\label{fig:ablation_command}
\end{figure}

\paragraph{Ablation command}
Figure~\ref{fig:ablation_command} shows that when choosing the command for exploration there is a slight advantage for asking the generator for a policy whose return exceeds the best return so far by 20. However, just asking for the maximum return (drive parameter = 0) is also competitive.

\section{Environment details}
\label{ap:env_details}

MuJoCo~\citep{todorov} is licensed under Apache 2.0.

\label{ap:ap}

\end{document}